\documentclass[letterpaper]{article} 
\usepackage{aaai23}  
\usepackage{times}  
\usepackage{helvet}  
\usepackage{courier}  
\usepackage[hyphens]{url}  
\usepackage{graphicx} 
\usepackage{natbib}  
\usepackage{caption} 
\usepackage{algorithm}
\usepackage{algorithmic}

\usepackage{newfloat}
\usepackage{listings}
\DeclareCaptionStyle{ruled}{labelfont=normalfont,labelsep=colon,strut=off} 
\floatstyle{ruled}
\newfloat{listing}{tb}{lst}{}
\floatname{listing}{Listing}

\copyrighttext{Presented at the AI-HRI Symposium at AAAI Fall Symposium Series (FSS) 2022}

\title{AI-HRI Brings New Dimensions to Human-Aware Design for Human-Aware AI}
\author {
  Richard G. Freedman
}
\affiliations{
    SIFT\\
    rfreedman@sift.net
}

\usepackage{amsmath}
\usepackage{amssymb}

\begin{document}

\maketitle

\begin{abstract}
Since the first AI-HRI held at the 2014 AAAI Fall Symposium Series, a lot of the presented research and discussions have emphasized how artificial intelligence (AI) developments can benefit human-robot interaction (HRI).  This portrays HRI as an application, a source of domain-specific problems to solve, to the AI community.  Likewise, this portrays AI as a tool, a source of solutions available for relevant problems, to the HRI community.  However, members of the AI-HRI research community will point out that the relationship has a deeper synergy than matchmaking problems and solutions---there are insights from each field that impact how the other one thinks about the world and performs scientific research.  There is no greater opportunity for sharing perspectives at the moment than human-aware AI, which studies how to account for the fact that people are more than a source of data or part of an algorithm.  We will explore how AI-HRI can change the way researchers think about human-aware AI, from observation through validation, to make even the algorithmic design process human-aware.
\end{abstract}

\section{Introduction\label{sec:introduction}}
The past seven Artificial Intelligence for Human-Robot Interaction (AI-HRI) Symposia weave a story about searching for identity and establishing community.  Between the themes and goals each year, one recurring topic has been how researchers in each of the AI and HRI fields can contribute to problems that matter at their integration, where robots employ AI algorithms that facilitate their interactions with people.

The framing of this definition of AI-HRI has generally (but not always) separated the high-level roles researchers specializing in one of the two fields play.  HRI-focused members of the AI-HRI community identify problems that could benefit from AI and design experiments (in controlled environments or the wild) that examine interactions between people and robots employing existing AI algorithms.  AI-focused members of the AI-HRI community develop algorithms and models that solve these presented problems as well as think about how people reason and learn in embodied interaction tasks.  This presents a bidirectional synergy where everyone in the AI-HRI community can help each other through an iterative progression of improvement, from identification and specification of a HRI problem, to formulation and repurposing a potential AI solution, to evaluation of the AI solution in the HRI context, to revealing new AI challenges within the HRI context through the results.  However, this process is a bit disjoint at present unless there is an established collaboration; most works presented at AI-HRI independently fit into one of these steps without a guarantee that other researchers in the community will adopt that work into the next step.  While this sounds like the community does not interact with one-another, the symposia have been very productive and collaborative through sharing knowledge and thinking about the future of the field as a whole (challenges, approaches, applications, ethics, etc.).

It is because these communities learn from each other and share their unique perspectives that AI-HRI differs from simply being two separate communities that trade problems and solutions, and this is what makes it the perfect community to explore the emerging trend for human-aware AI.  Years prior to the term's coinage \cite{ijcai16_specialTrack}, researchers in the AI-HRI community were identifying problems that needed human-aware AI and simultaneously working towards human-aware AI solutions \cite{robotConsidersHuman}.  These include studying which people to recruit for assistance in human-in-the-loop tasks \cite{Rosenthal:2012:MRP:2900929.2901020_CobotAskHelp}, legibility in motion planning \cite{Dragan:2014:IHO:2678082.2678168_disambiguousReaching}, responsive planning for closed-loop interaction \cite{fz_aaai17,ffgz_aihri2019}, and model reconciliation for generating explanations \cite{cszk_ijcai17}.  
While it is clear that the AI-HRI community benefits from and contributes to human-aware AI research, it is worth noting that the AI commuity at large has begun to approach human-aware AI research without as deep an understanding about the human aspect.

The AI-HRI community understands that people are neither machines, algorithms, nor exclusively sources of data, and this motivates thinking about humans not just for the sake of making AI human-aware.  Instead, the AI-HRI community understands the need for \emph{human-aware design} of human-aware AI, and we should consider how to formalize and communicate these practices for others to create effective human-aware AI systems.  These can benefit not just future AI-HRI research, but the experiences of people who will eventually interact with systems employing human-aware AI in the real world.  We introduce several directions in which the AI-HRI community can study, discuss, and develop human-aware design for human-aware AI.

\section{Defining Human-Aware AI\label{sec:whatIsHumanAware}}
After creating a track for human-aware AI at the IJCAI 2016 conference \cite{ijcai16_specialTrack}, \citeauthor{aaai18_invitedTalk} \shortcite{aaai18_invitedTalk} gave an invited talk at AAAI 2018 and later prepared a write-up for AI Magazine \cite{Kambhampati_2020_aimag} introducing the topic.  None of these present a formal definition for human-aware AI, but instead present motivations for what human-aware AI can achieve as well as provide some examples.  \citeauthor{freedman-murc23} \shortcite{freedman-murc23} incorporated human-aware AI into the 2023 EAAI Mentored Undergraduate Research Challenge, but also loosely defined it as AI that thinks about people and provided some examples of what is and is not human-aware AI.  These vague descriptions leave room for interpretation, but perhaps the AI-HRI community can help establish something more concrete.

To begin, what does it mean for robots to ``think about a person?''  For example, \citeauthor{DBLP:conf/atal/Kaminka13} posited that robots lack social intelligence without extensive domain-specific programming and proposed a list of tasks to address this issue \cite{DBLP:conf/atal/Kaminka13}.  Among the list of challenges to the AAMAS community, \citeauthor{DBLP:conf/atal/Kaminka13} includes ``fundamental mechanisms for understanding others,'' which gathers information (via communication and/or observation) about an agent to predict its actions and intents.  This alludes to plan, activity, and intent recogition \cite{sukthankar2014pair}, but there are a number of additional elements to consider such as when it is best to gather the information, by what means it is appropriate to gather that information, and how much information the robot needs to gather without invading privacy.  All of these tasks require person-specific considerations to establish social context \cite{briggs_aihri16}, assess factors like cognitive load \cite{lohan_cognitiveLoad_aihri18} to determine when some form of interaction is appropriate \cite{wilson_aihri2020}, and maintain trust \cite{lee_aihri2020}.  Do we deem each of these indepenent instances of human-aware AI, or should we view the entire system as a singular instance of human-aware AI?  The answers to such questions will influence the ways we think about human-aware design because it impacts the identification of which part(s) of the robot's AI can become human-aware as well as the properties that make it possible (see Section~\ref{sec:makeHumanAware}).

Another possible layer of complication to consider when defining human-aware AI is whether machine learning can extrapolate human-awareness from data. 
In many ways, this is analogous to Searle's Chinese Room Argument \cite{searle_1980}: if a person observes just the input stimuli and output responses of some human-aware process, then are patterns that emulate this mapping aware of the human as well?  Would some form of input data describing the people be sufficient to mark a learned model as human-aware?  Declaring these precedents gives the AI-HRI community a starting point on top of which we can develop assessments and evaluations (see Section~\ref{sec:humanAwarenessTests}). 

\section{What Should Be Human-Aware, and How Much?\label{sec:makeHumanAware}}
AI systems for robots that interact with people can become very complex because the architecture must support multiple sensing modalities and capabilities.  For example, the EU-funded MuMMER project's robot designed to naturally interact with people in public spaces has audio and visual sensing to work on seven tasks (signal processing, state estimation, signal generation, conversation, route guidance, motion planning, and shared visual perspective planner) \cite{mummer_aihri2019}.  Although it may be possible to make all the components in a single AI-HRI system human-aware, it is important to consider whether we should and why.  From a human-aware design perspective, which AI systems within the architecture actually think about the human?  When two standalone AI systems are connected in the robot's framework, how do they communicate and share information properly if one is human-aware and the other is not?  Are there additional ways that the robot's AI should be human-aware holistically that any single AI system within the architecture cannot accomplish?

For individual AI systems, one element of human-aware design to consider is where the algorithms, models, and data can be modular to enable personalization.  It is common to use off-the-shelf methods when they are available, and it is easier to 
make a ``one-size-fits-all'' AI system.  Together, this leads to robots using AI systems that, even if human-aware, are designed with an 
assumed 
user in mind or at risk of dataset bias.  For example, a recommender system that makes suggestions based on how a person's recent choices match other people's choices assumes a set of selections has a common relationship that all users share.  This does not consider \emph{why} the person made their choices, such as 
exploring something new, having a personal relation outside the 
one that rises from the assumption, and individual preferences.

The AI-HRI community has considered human preferences in the form of ordering constraints for reasoning \cite{wilson_preferences}, means of action selection that match user expectations \cite{marge_preferences}, objective functions for ideal criteria to satisfy when possible \cite{gombolay_preferences}, and information needed for an explanation to be relevant \cite{userCenteredExplanations_aihri21}.  Each representation of preferences was selected to account for things that matter to the person while the robot performs the specific intelligence task, and it is important to document why this unique representation was chosen and how human-aware AI uses it in the algorithm.

Outside of preferences, there are other ways to personalize that account for the person's mental model, cultural background, and/or physical stature---not everyone is accounted for when human-aware AI is designed under the ``one-size-fits-all'' approach.  It is a habit in many AI algorithms, not just machine learning, to have a number of tunable parameters that adjust how the algorithm performs and revise the output.  We frequently assign parameter values that yield the best performance for our experiments, but what if different instances need different parameter values?  When is dynamically overfitting to each subpopulation actually adjusting behavior per person who interacts with the robot?  Responsive planning considers a parameter for how suboptimal a person is at solving a task \cite{fz_aaai17} so that a robot does not dismiss the true goal if that person makes a mistake or generally struggles with the tasks at hand.  Likewise, model reconciliation \cite{cszk_ijcai17} searches for a mapping of the problem and domain spaces that fit a user's constraints even if they are incorrect, which enables the robot to put itself in their shoes and establish common ground.  There is a growing realization that training a model on data is not sufficient when that data fails to represent a group of users or captures historical biases \cite{rudin_nmi19}, and AI-HRI is no exception when we consider how common speech is used as a means of communication.  Dialects, accents, and even word choice are not represented equally in speech-to-text systems \cite{speechToText_disparities}; human-aware design can address this through creating methods to identify where the representation gaps are and collect data that can fill those gaps.  With the extra data, a seemingly non-human-aware AI approach like text-to-speech can become human-aware by determining which language model to use for facilitating communication during its interactions.





It is important to address ethical concerns that arise from human-aware design.  Regardless of whether researchers intentionally apply human-aware design to their human-aware AI, the researcher is still 
making decisions about the input assumptions, collected data, algorithmic control flow, output format, and more.  The AI-HRI community is not immune to this even if we carefully think about the human-aware design because we are individuals with our own experiences that are not the same as everyone else's.  The National Human Genome Research Institute's recent strategic vision acknowledges the need for a diverse genomics workforce in order to increase the number of unique experiences and skillsets that everyone can provide \cite{nhgri-improve-diversity}, and this generally applies when thinking about the potential people who might interact with a robot in the near future.  
If we simply leave it up to the researchers to make decisions about human-aware AI, then what could our robots be missing in future interactions?  Besides diversity among researchers, we also need to think about diversity outside of researchers because the robots will interact with a variety of users. 
How do we get access to these inputs?  Is it enough to run Wizard-of-Oz studies \cite{woz_10.5898/JHRI.1.1.Riek} with a diverse population of recruited subjects?  Do we need to educate non-researchers more about AI \cite{aiEducation_bluesky} and HRI \cite{reframingHRIeducation} so that they can provide informed insights and feedback?  It is important to remember that if human-aware AI for HRI must think about people, then human-aware design requires us to think about people 
beyond
how their existence affects the 
technology we create.

\section{Evaluating AI for Human-Awareness\label{sec:humanAwarenessTests}}
Some of the most important steps in the scientific method are experimentation and analysis of the results.  Both the AI \cite{makeAIreproducible} and HRI \cite{makeHRIreproducible} research communities often struggle with reproducibility, but there are a number of challenges and competitions that introduce benchmarks, metrics, and tests for related works to evaluate themselves under similar conditions.  Although reproducible research in human-aware AI is strongly encouraged for good scientific practice, we first need to establish how to evaluate human-aware AI research under our definitions (see Section~\ref{sec:whatIsHumanAware}) in order to have studies to reproduce.

Considering the popularity of tests to challenge the research community to achieve some milestone, such as the classic Turing Test \cite{turingTest} and more recent Lovelace 2.0 Test \cite{riedl:aaai-turing2015} for creative intelligence, is there a test for human-aware intelligence?  If there is, then it will not be as simple as checking a list of criteria.  As an example, consider the following condition based on the oversimplified belief that human-aware AI function $f$ will not give the same response to everyone because it can think about them as individuals beyond the function parameters ($H$ is the set of all humans and $context\left(\cdot\right)$ yields relevant information about the world/situation with respect to $\cdot$, such as their perspective, entities around them, related history, etc.):
\begin{align}\label{eq:badHAAItest}
  \begin{split}
    \exists h_{1}, h_{2} \in H . & f\left(x_{1}^{h_{1}}, \ldots, x_{k}^{h_{1}}, context\left(h_{1}\right), h_{1}\right) \neq \\
     & f\left(x_{1}^{h_{2}}, \ldots, x_{k}^{h_{2}}, context\left(h_{2}\right), h_{2}\right) \\
  \end{split}
\end{align}
when $\forall i \in \left\lbrace 1, \ldots, k \right\rbrace . x_{i}^{h_{1}} = x_{i}^{h_{2}}$.  It is trivial to prove that this test is insufficient for validating human-awareness: \textit{let function $f$ be a random number generator; there is a decent probability that Equation~\ref{eq:badHAAItest} evaluates to true, but selecting a number at random clearly never thinks about the human or even uses any of the function's parameters} $\Box$.  Although it is more hypothetical, we can also present a convincing example that this test is unnecessary for validating human-awareness: \textit{let $f$ be an algorithm that spends a lot of computational power pondering about people on a tough topic (like violating Asimov's Laws) and, no matter how long $f$ thinks about each person, it fails to find a reason to change its answer (cannot harm this person despite everything because...).  $f$ is likely human-aware, but Equation~\ref{eq:badHAAItest} evaluates to false}.  The latter discussion provides evidence that what goes on under-the-hood of $f$ matters as much as the inputs/outputs, further supporting the need for interpretable models and explainable methods.  The aforementioned tests only need to inspect the inputs and outputs because the expressed behavior or deliverable creation is the point of evaluation.

From a human-aware design perspective, is there a metric that can measure how human-aware an AI system or robot with an architecture of multiple human-aware AI systems is?  Would such an assessment be quantitative or qualitative?  There is increasing evidence that most people think qualitatively \cite{numbersNeedWords} in domain-specific contexts \cite{domainSpecificThinking} rather than quantitatively in domain-independent contexts, and the latter resembles a computer program.  At Stuart Russell's invited talk at AAAI 2020, Barbara Grosz reminds us in her post-presentation question that there are various models for cooperation that do not rely on maximizing numerical utilities and that humans are not necessarily the sum of individuals \cite{russellAAAItalk2020}.  If we measure human-awareness as an accuracy percentage like traditional AI assessments on benchmarks, can people (not just AI-HRI researchers) confidently determine how well a robot is able to think about them and/or other people?  Are distributions over Likert scale survey results from people \emph{with privilege to have access to participation in AI-HRI studies} enough to determine how well a robot is able to think about other people?  \citeauthor{likertAssess_aihri16} \shortcite{likertAssess_aihri16} presented at AI-HRI 2016 that we need to be careful with how we statistically evaluate Likert scale items as well.  Given the balance of pros and cons, 
we challenge the AI-HRI community to think about whether a middle-ground exists that correlates objective and subjective evaluations where both qualitative and quantitative evaluations of human-aware AI have useful meaning.



As a final direction to consider with respect to evaluation of human-aware AI, we propose that human-aware design means we also need to think about the research and development of AI more like how we think about the research and development of usable software and hardware: iterative research and evaluation with human users.  Traditionally, the HRI research community (as well as the HCI and UX communities) runs focus groups, interviews, and/or Wizard-of-Oz (WoZ) studies in order to collect information about what people expect prior to doing any research and development.  In contrast, the AI research community frequently completes their initial AI system before considering any tests, which are generally computing metrics and comparing results on the benchmarks.  Due to this, AI-HRI research that tests how people respond to robots using existing AI systems often has to wait until the AI is completely implemented to test it as well.  Letting people interact with the AI in human subjects research throughout development allows us to find pitfalls that might not appear in datasets and benchmarks; it could even identify new research problems based on how the interaction experience plays out, introducing new questions about human-awareness \cite{ffgz_aihri2019}.  This claim is not likely going to surprise members of the AI-HRI research community, but it is important to keep in mind that AI researchers who are starting to work on human-aware AI are not likely to consider this.  Not considering the involvement of people throughout the design of human-aware AI can lead to many of the concerns we have discussed throughout this paper, and it is our responsibility to share good practices from our own experiences so that new researchers, whether in AI-HRI or not, have the information they need to design human-aware AI systems that we feel safe and comfortable deploying on our robots.

This motivation does not answer the question about how to perform iterative research and testing, though.  What comes in between the original focus groups/interviews/WoZ studies and the human-robot laboratory trials?  Interactive architectures composed of many AI systems are rarely complete enough to deploy in a laboratory setting until they are finished, but it would benefit from iterative evaluation along-the-way to update the human-aware AI capabilities when problems first appear rather than at the end.
WoZ studies let us identify what people do in various situations to prepare a checklist of things to think about when creating human-aware AI.  The key is that there are two ways to learn from people in a WoZ study: the human interacting with the robot reveals how people treat the robot and their expectations of what the robot can do, but the robot's actor in another room gives us insights into what the robot should do if it was a human addressing the problem.  The actor provides us with a unique opportunity for a before/after assessment with respect to each capability of the human-aware AI system.  In contrast to learning from demonstration \cite{lfd_aihri14} where the actor's actions serve as training data, these responses can serve as a benchmark to deteremine when certain capabilities \emph{seem} ready to think about people in interactions.  In order to validate this off-line assessment and determine how people change their interactions (the human-aware AI is rarely going to be exactly like the actor, and these differences can affect the interactive experience), it is important to run another WoZ-like study where we \emph{replace part of the actor's control with the corresponding human-aware AI system}.  Similar to cyborg chess where humans use machines as a guide and tool while playing chess \cite{cyborgChess}, \emph{cyborg WoZ} studies still need humans to puppet the robot.  However, the AI system(s) will do whatever they can to inform, guide, or instruct the human; that is, the actor turns over some of their autonomy to the AI.  We encourage the AI-HRI community to develop and test additional techniques, but we introduce cyborg WoZ studies as a preliminary approach to iteratively apply human-aware design throughout the research of human-aware AI.

There are some ethical issues we need to consider about cyborg WoZ studies, and the AI-HRI community will likely need to consider these for future iterative evaluation techniques as well.  What if the human-aware AI instructs the actor to do something bad (out of error, rather than malice), which poses a case of the Milgram experiment on obedience \cite{milgram1974obedience} if we instruct the actor to follow the AI?  In contrast, what if the human actor disagrees with the human-aware AI's observation and advice, whether by bias or access to different information, and chooses to act on their accord?  Would either of these situations violate the assessment of the robot's human-awareness because the human and machine deviated from each other?  It might be the case that some of these issues relate to whether the AI is providing information (perception, assessment of situation) or making any decisions (planning, policy execution), 
but it is important that we continue to think about all the people involved in experiments throughout the human-aware design process. 

\section{Conclusion\label{sec:conclusion}}
The AI-HRI community brings many unique perspectives and practices due to integrating two fields that have relatively different approaches to doing research.  Although human-aware AI is getting recognized in the AI community as technologies like mobile devices become more ubiquitous, it does not mean that their approaches will consider the human's role as thoroughly.  People are more than a part of the algorithm, and we hope the AI-HRI community will help others understand and learn to think about human-aware design.  It can lead to many new methods for human-aware AI in HRI, mututally benefitting everyone.  
As a start, we proposed some definition, implementation, and evaluation considerations. 
  We encourage AI-HRI researchers to continue this discussion, discover new approaches, and continue to share their scientific insights.

\section{Acknowledgments}
The author thanks the anonymous reviewers for their feedback to improve the quality of this paper.  The author also thanks Todd W.~Neller, Vasanth Sarathy, Sonja Schmer-Galunder, and Helen Wauck for engaging discussions about various topics that inspired some of the content in this paper.  Nathaniel Budijono's, Jack Maloney's, and Ratul R. Pradhan's time listening to the author's various attempts at defining human aware-AI and sharing their own interpretations is greatly appreciated.

\bibliography{bluesky_aihri}

\end{document}